# Enhancing efficiency of object recognition in different categorization levels by reinforcement learning in modular spiking neural networks

Fatemeh Sharifizadeh[a], Mohammad Ganjtabesh[a,*], Abbas Nowzari-Dalini[a]

[a]*Department of Computer Science, School of Mathematics, Statistics, and Computer Science, University of Tehran, Tehran, Iran.*

## Abstract

The human visual system contains a hierarchical sequence of modules that take part in visual perception at superordinate, basic, and subordinate categorization levels. During the last decades, various computational models have been proposed to mimic the hierarchical feed-forward processing of visual cortex, but many critical characteristics of the visual system, such actual neural processing and learning mechanisms, are ignored. Pursuing the line of biological inspiration, we propose a computational model for object recognition in different categorization levels, in which a spiking neural network equipped with the reinforcement learning rule is used as a module at each categorization level. Each module solves the object recognition problem at each categorization level, solely based on the earliest spike of class-specific neurons at its last layer, without using any external classifier. According to the required information at each categorization level, the relevant band-pass filtered images are utilized. The performance of our proposed model is evaluated by various appraisal criteria with three benchmark datasets and significant improvement in recognition accuracy of our proposed model is achieved in all experiments.

## 1. Introduction

Object recognition can take place at different levels of hierarchical abstraction. Psychologists have commonly divided this abstraction into three levels: superordinate (e.g., animal vs. non-animal), basic (e.g., bird vs. mammal), and subordinate (e.g., pigeon vs. duck) [1]. Some studies found evidence that these identification terms lay within a taxonomy of levels from the superordinate to subordinate [2, 3, 4, 5, 6].

The biological evidence reveals that the visual input is processed from the coarse-grained features (low spatial frequencies) to fine-grained features (high spatial frequencies) [7, 8]. In general, larger, global and low spatial frequency (LSF) elements of the input are processed with a higher priority than the small, local and high spatial frequency (HSF) elements [9, 10], especially when the different levels of categorization are involved. The results of some psychophysics experiments and computational models showed that the superordinate level categorization mainly relies on LSFs and the accuracy of basic and subordinate levels are enhanced by increasing the spatial frequency [3, 4, 11].

During the last decades, various computational models examined each categorization level individually. These models have been proposed to mimic the hierarchical feed-forward processing of the object recognition task in a bottom-up cascade of human cortical regions [12, 13, 14, 15, 16, 17, 18, 19, 20, 21, 22, 23, 24, 25]. Despite the architectures of these models are inspired by the human visual cortex (a hierarchy of layers with gradually increasing receptive fields), the search for brain-inspired computational models continues and is attracting more and more researchers from around the world. In general, the research challenges of the object recognition task at different categorization levels are mainly related to the following five issues. First, till now, no computational model has been proposed to perform the recognition task at all three categorization levels, i.e. the existing computational models deal with each categorization level individually. Second, the complexity of object recognition is increased by moving from the superordinate to subordinate level. Third, most of the existing computational models are not really biologically plausible because the actual neural processing and learning mechanisms of the visual cortex are neglected in these models. Forth, during many object recognition tasks, the input may be ambiguous, i.e. appear with different variations, noise or occlusion, and hence, only partial and limited information about the objects is accessed which hampering an explicit mapping from input to object category.

With proceeding in the levels of categorization, the complexity of object recognition is gradually augmented and fine discrimination between similar objects becomes harder [26], especially at the subordinate level. Recognizing finegrained categories at the subordinate level is quite challenging because the visual differences between the fine-grained categories are small and can be easily overwhelmed by the variation of the objects. The difficulties of this task are due to the discriminative region localization and fine-grained feature learning [27, 28]. Consequently, human-defined regions or the regions learned by existing methods may not be optimal for machine classification and also, learning of subtle visual differences existed in local regions from similar fine-grained categories is difficult.

The computing units in most of the computational models, such as Deep Convolutional Neural Network (DCNN), send floating-point values to each other which correspond to their activation level, while, biological neurons communicate to each other by sending spikes. Neurons fire a spike only when they must transmit an important message, and some information can be encoded in their spike times. Inspired by biologically-realistic models of neurons, spiking neural networks (SNN) have been developed to carry out the computation. Many researchers studied the configuring parameters of SNN, such as the number of layers (shallow or deep) [29, 30, 31], types of connectivity schemes (recurrent, convolutional, and fully connected) [29, 32, 33, 34], and information encoding (rate-based and temporal coding) [29, 30, 34, 35]. Different learning techniques are also applied to SNNs, from backpropagation [33, 36], tempotron [30, 37], and other supervised techniques [38, 35], to unsupervised Spike Timing-Dependent Plasticity (STDP) and its variants [34, 39]. Regarding the synaptic plasticity in different brain areas, STDP-based SNNs are the most biologically plausible ones. STDP learning rule works by considering the firing time difference between pre- and post-synaptic neurons. According to STDP, if the pre-synaptic neuron fires earlier (later) than the post-synaptic one, the synapse is strengthened (weakened). STDP works well in finding statistically frequent features, however, as any unsupervised learning rule, it faces difficulties in decision making and hence, the external classifiers are usually required. So, several studies suggest using reinforcement learning (RL) which has a major impact on decision making and behavior formation. In the RL, the learner is encouraged to repeat rewarding behaviors and avoid those leading to punishments [40, 41, 42]. The RL is inspired by the brain's reward system and it is shown that dopamine is an organic chemical substance involved in reward-motivated behavior [41] and it affects the synaptic plasticity, such as changing the polarity [43] or adjusting the time window of STDP [44, 45]. While unsupervised learning is one of the driving forces of plasticity, STDP neglects the information regarding reward and punishment. Thus, several studies have modeled the role of the reward system by modulating the weight change determined by STDP, which is called reward-modulated STDP (R-STDP) [46, 47, 48]. R-STDP stores trace of eligible synapses for STDP and apply the modulated weight changes at the time of receiving a modulatory reward or punishment signal.

In this paper, we introduce a new computational model that incorporates the reinforcement learning in the bottom-up processing of the input image for object recognition in different categorization levels. Our proposed model is comprised of three modules corresponding to the three levels of categorization. In the proposed model, the LSF elements of input image, with high priority, are entered into the superordinate module and then a fast decision is made at this level. Because of extracting coarse-grained features, low number of classes, and simplicity of classification at this level, the decision of this module is quickly determined. Pursuing the line of biological inspiration, we used a spiking neural network equipped with reinforcement learning rule as a module at each categorization level, where R-STDP is used as a learning rule. Each module solves the object recognition problem at each categorization level, solely based on the earliest spike of class-specific neurons at its last layer, without using any external classifier. It is expected that the proposed model performs well in confronting with complex objects such as occluded or ambiguous input images. The ETH-80 [49], a subset of CU3D [17], and a subset of ImageNet [50] datasets are utilized to evaluate our model. The results of the experiments, under different conditions, show the efficiency of our proposed model.

The rest of this paper is organized as follows. In Section 2, the idea for developing our proposed model as well as its details are described. Section 3 provides and illustrates the experimental results. Finally, the paper is summarized and concluded in Section 4.

## 2. Materials and methods

In this paper, we propose a novel computational model, for object recognition in different categorization levels, namely superordinate, basic, and subordinate, that mimic the hierarchical processing in the visual cortex. For each categorization level, a corresponding module is employed in the model. Each module comprises a hierarchical spiking neural network equipped with R-STDP learning rule, entitled R-SNN, to solve the object recognition task in that categorization level. The recognition in each module is done based on the neural activity of its last layer, without using any external classifier.

By using all frequency bands of the input image at each categorization level, extra and redundant information is fed into the modules. Hence, the role of applying relevant information for each level is fading out. Therefore, according to the required information at each categorization level, the relevant band-pass filtered images are utilized. More specifically, the information of low, intermediate, and high spatial frequencies (LSF, ISF, and HSF) of the input images are employed in the first, second and third module of proposed model, respectively.

### 2.1. The overall structure of modules

The schematic representation of the architecture of each module with two simple and two complex layers that are alternately arranged, is shown in Figure 1. The parameters of each module should be adapted for the desired recognition task.

The first simple layer (S1), extracts oriented edges from the frequency-filtered input image and turn them into spike latencies using intensity-to-latency encoding. To this end, the input is convolved with Gabor filters of four different orientations and hence, four feature maps are achieved in this layer, each represents the edges in a specific orientation. For each feature map, a 2-D lattice with the same size is placed that contains dummy neurons for spikes propagation. The obtained feature maps are converted to the spike latencies, which are inversely proportional to the saliency of the edges, i.e., the more salient the edge, the earlier the corresponding spike is propagated. The spikes are sorted in ascending order by their latencies and propagated sequentially (i.e. the first spike is propagated in time step $t = 1$, the second one in $t = 2$, etc).

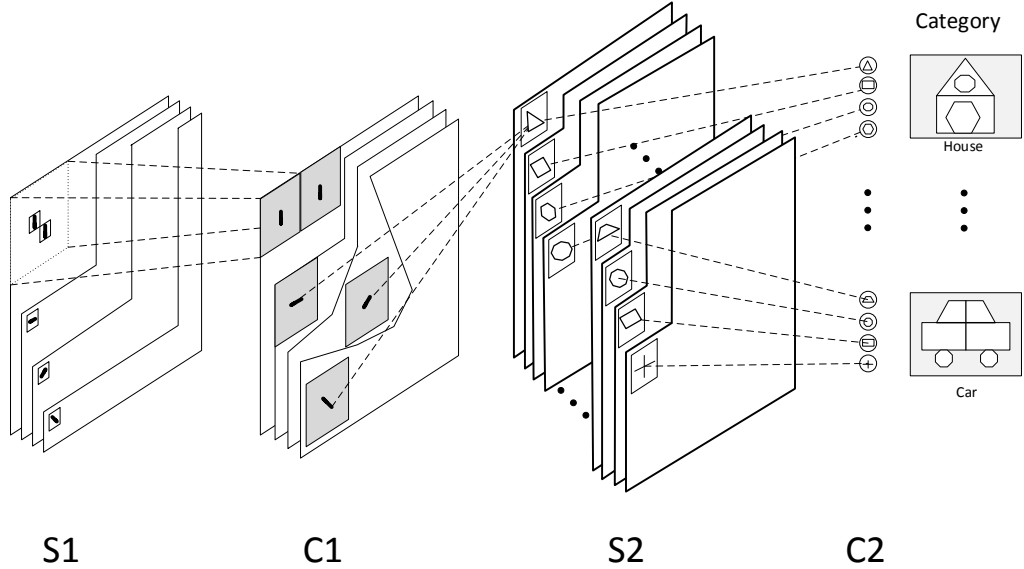

**Figure 1:** The structure of the R-SNN module with four layers. The edges of the input image are extracted in the S1 layer. A local max-pooling operation is implemented in the C1 layer and the spikes are propagated into the next layer. In layer S2, there are 2-D lattices of IF neurons, that learn to extract complex features using the R-STDP learning rule. The final decision of the R-SNN module is determined by the neurons in layer C2 that are associated with the class labels.

The second layer (C1) performs local pooling over the spikes coming from S1. In C1 layer, four 2-D neuronal lattices corresponding to each of the orientations is placed. Each neuron in C1 executes a local pooling operation over a window of size $w_{c1} \times w_{c1}$ and stride s$c1$ ($s_{c1} = w_{c1} - 1$) on S1 neurons in the corresponding lattice, after which, it emits a spike immediately after receiving its earliest input spike. The pooling operation decreases the number of required neurons and shrinks the existed redundancy in layer S1, hence increasing computational efficiency. In addition, a local invariance to the position of oriented edges occurs. Also, to propagate more salient information two kinds of lateral inhibition mechanism are applied. If one neuron fires from the i[th] lattice at a particular position, 1) it prevents the other lattices' neurons at the same position to fire, and 2) the latencies of adjacent neurons in the same lattice are increased with respect to their Euclidean distance from the fired neuron.

The third Layer (S2) combines the incoming information from C1 about the oriented edges and turns them into more complex features. This layer comprises n 2-D lattices of integrate and fire (IF) neurons with threshold θ. These neurons have no leakage and are allowed to fire at most once while an image is being presented. Each S2 neuron collects its inputs from a window of size $w_{s2} \times w_{s2} \times 4$ of C1 neurons via plastic synapses. To detect a specific feature over the entire spatial positions, the weight sharing mechanism is employed for the neurons in the same lattice. The initial weights of the synapses are randomly generated. The membrane potential of i[th] neuron in S2 at time step *t* is updated by the following equation:

$$V_i(t) = V_i(t-1) + \sum_{j \in P(i)} W_{i,j} \times \delta(t - t_j) \qquad (1)$$

where *P(i)* is the pre-synaptic neurons of i[th] neuron, $t_j$ is the firing time of the j[th] neuron at C1, δ is the Kronecker delta function, and $W_{i,j}$ defines the synaptic weight between neurons *i* and *j*. When S2 neurons fire, their synaptic weights are updated according to the order of pre- and post-synaptic spikes, as well as the reward/punishment signal. This signal is derived from the activity of the next layer, that also indicates the module's decision.

The fourth layer (C2) determines the final decision of the module. This decision is used to generate a reward/punishment signal, which modulates the synaptic plasticity of S2 neurons. This layer contains exactly n neurons, one neuron for each lattice in the S2 layer, and each neuron performs a global pooling operation. A C2 neuron only propagates the first received spike from its corresponding lattice. The firing time of $i^{th}$ (i=1,...,n) neuron at C2 layer is computed as follows:

$$t_i = \min_{j \in P(i)} \{t_j\} \qquad (2)$$

where P(i) denotes S2 neurons in the $i^{th}$ neuronal lattice (pre-synaptic neurons), and $t_j$ is the firing time of the $j^{th}$ neuron at S2 layer. As mentioned before, the activity of C2 neurons indicates the module's decision. So, C2 neurons are divided into several groups and each group is assigned to a particular class of input image in that module. In this way, a group that propagates the earliest spike among the C2 groups specifies the module's decision. Let n denotes the number of S2 lattices and m denotes the number of input classes. Then the module's decision (D) for each input image is calculated by:

$$D = g(F), \quad F = \min_{1 \leq i \leq n} \{t_i\} \qquad (3)$$

where $F$ denotes the index of a C2 neuron which fires first, $t_i$ is the firing time of the $i^{th}$ neuron at C2 layer, and $g: \{1, ..., n\} \rightarrow \{1, ..., m\}$ is a function that returns the group index of a C2 neuron. If D matches (does not match) to the correct category of the input image, the module receives a reward (punishment) signal. When more than one neuron fire, the one with minimum spike time is selected. Moreover, when all the C2 neurons are silent, the reward/punishment signal is not generated and hence, the weights are not changed.

## 2.2. The learning mechanism of modules

In order to train the modules, the weights of S2 neurons are updated using the R-STDP learning rule. According to the correctness/incorrectness of the module's decision, the magnitude and/or the polarity of weight change are tuned by the reward/punishment signal. One-winner-takes-all learning competition among the S2 neurons is also employed, in which the neuron with the earliest spike is the winner and it is eligible to update the synaptic weights, that are shared among the other neurons at the same lattice. In fact, the winner neuron determines the module's decision. According to the R-STDP, the amount of weight change ($\Delta W_{i,j}$) for the synaptic connection between two neurons $i, j$ is calculated by:

$$\Delta W_{i,j} = \Gamma_{i,j}.W_{i,j}.(1 - W_{i,j}); \quad \Gamma_{i,j} = \begin{cases} R.A_r.M_r^+ + P.A_p.M_p^- & : if \ t_j - t_i \leq 0 \\ R.A_r.M_r^- + P.A_p.M_p^+ & : if \ t_j - t_i > 0 \end{cases} \qquad (4)$$

where the values of $M_r^+$, $M_r^-$, $M_p^+$, $M_p^-$ scale the magnitude of weight change ($M_r^+$, $M_p^+$>0 and $M_r^-$, $M_p^-$<0). The values R and P depend on the generated reward/punishment signal: if reward signal is generated, then R = 1 and P = 0; whereas if punishment signal is generated, then R = 0 and P = 1. In the reinforcement learning, the overfitting problem usually happens due to the unbalanced impact of reward and punishment in high rate of correct-classified and mis-classified samples. In the high rate of correct-classification cases, the rate of reward acquisition increases and the module prefers to exclude misclassified samples by getting more and more selectivity to correct ones and remain silent for the others. Similarly, in the high rate of mis-classification cases, the module receives more punishment signals, which rapidly weakens synaptic weights and generates dead or highly selective neurons that cover a small number of inputs. To cope with this difficulty, the adjustment factors ($A_r = N_{incorrect}/N$, $A_p = N_{correct}/N$) are applied in the weight modification, where N*correct* and N*incorrect* respectively denote the number of samples that are classified correctly and incorrectly over the last batch of N input samples. Therefore, the impact of correct and incorrect training samples is balanced over the trials. Also, the multiplication between the terms $\Gamma_{i,j}$ and $W_{i,j}.(1 - W_{i,j})$ leads to keep weights between range [0, 1] and stabilizes the weight changes as they converge.

## 2.3. The overall structure of the proposed model

The overall sketch of our proposed model is presented in Figure 2, in which the input images could be classified at all three categorization levels.

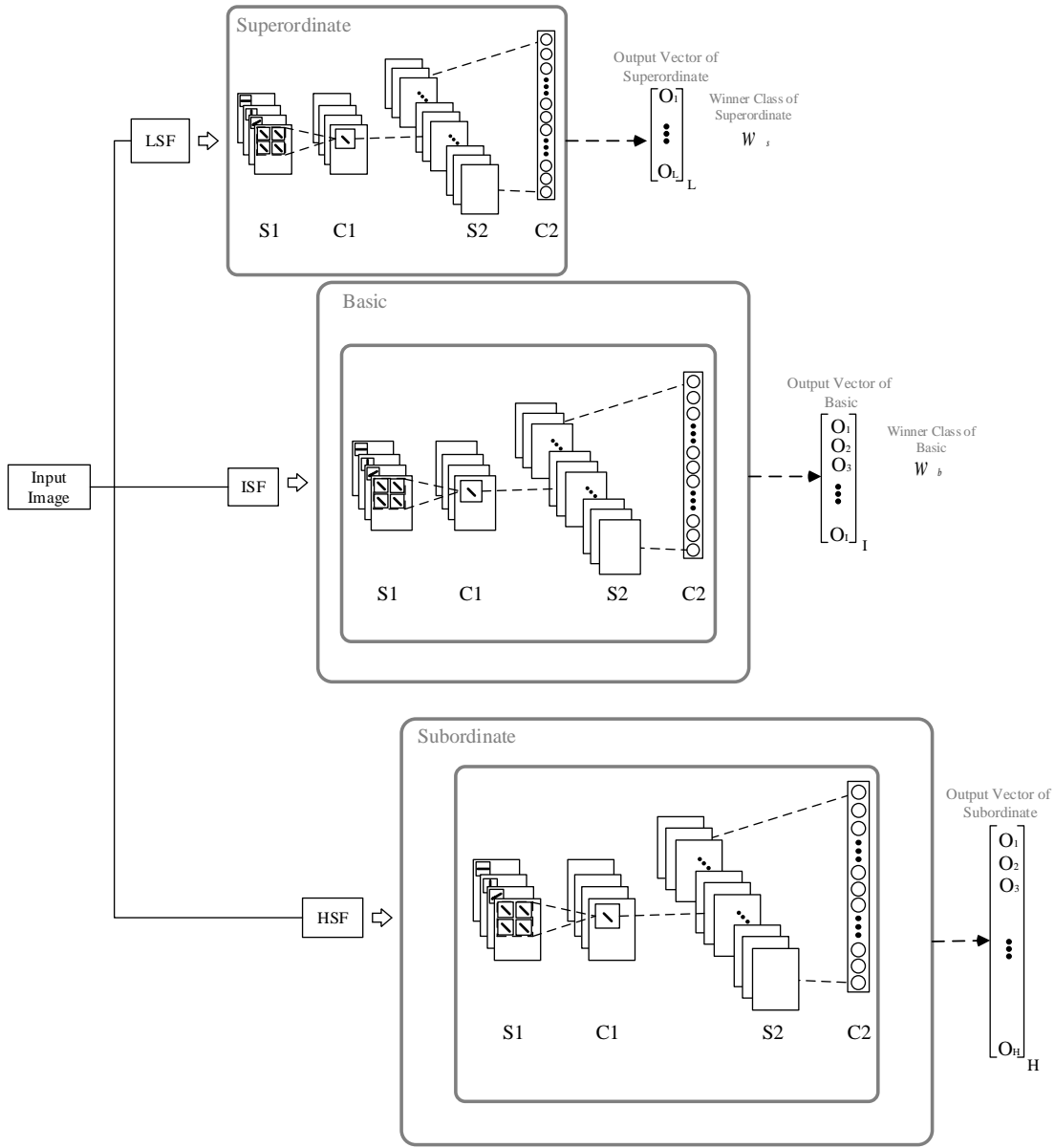

**Figure 2:** The sketch of our computational model in different categorization levels. The model has three R-SNN module for categorization in the superordinate, basic and subordinate levels, which are specified by gray rectangles. LSF, ISF, and HSF information of input is used in the first, second and third module, respectively. The output of each module is considered as output class at the corresponding categorization level.

The details of each categorization level of proposed model are as follows:

**Superordinate level:** The LSF elements of the input image are propagated to the first module, and the output of this module determines the class of input image at this level. Let *L* denotes the number of object classes at this level, so the number of neuronal groups in the C2 layer of this module is exactly *L*. Therefore, the output of this module is a vector of size *L*, which determines the winner class at this level.

**Basic level:** At this level, the ISF elements of the input image are propagated into the second module. As each class *c* of the superordinate level contains $ns_c$ species (categories) at the basic level, each image at the basic level is classified into I different classes. If there are *n* lattices in the S2 layer of the basic module, correspondingly there are *n* neurons in the C2 layer, each assigned to one of the S2 lattices, that are divided into *I* groups. As it is argued, with proceeding in the levels of categorization, the complexity and the number of classes are gradually increased, and hence, accurate analysis is required to perform the object classification task. Finally, the output vector of this module with *I* components is formed, where *I* indicates the number of classes at the basic level.

**Subordinate level:** The process of the third (subordinate) module is very similar to the second (basic) one. Since the number of classes is further increased and objects of different classes highly resemble each other at this level, more and finer features are needed for distinguishing them. Therefore, the HSF elements of the input image are propagated into the third module. Again, each class c of the basic level contains $ns_c$ species (categories) at the subordinate level, and each image at the subordinate level is classified into *H* different classes. If there are m lattices in the S2 layer of the subordinate module, correspondingly there are m neurons in the C2 layer, each assigned to one of the S2 lattices, that are divided into *H* groups. Each S2 lattice extracts its related class-specific features and the neuronal groups of C2 layer generate the final decision of this module. Finally, after performing the customary process in this module, the input image is classified at the subordinate level.

In the experimental results, it is demonstrated that reinforcement learning can play important roles in forming robust object recognition at different levels of categorization and also in the case of ambiguous object recognition.

## 3. Results

To evaluate the performance of our proposed model as well as to exhibit its advantages in comparison with the other methods, various experiments under different criteria are carried out. 3.1. Specifications of employed datasets The proposed model is evaluated with three benchmark datasets, namely ETH-80 [49], CU3D-100 [17], and ImageNet [50].

The ETH-80 dataset contains 80 3D objects in eight different object categories including apple, car, cow, cup, dog, horse, pear, and tomato. Each object category includes 10 exemplars and each exemplar is photographed in 41 viewpoints with different view angles and tilts. Three categorization levels for the employed subset of ETH-80 dataset are shown in Figure 3. The superordinate level contains the animal, object, and fruit categories. At the basic level, the animal category contains the dog, cow, and horse; the object category includes the cup and car; and the fruit category includes the pear, apple, and tomato. The subordinate level includes three exemplars of each category at the basic level, i.e., there is a total of 8×3=24 categories at this level.

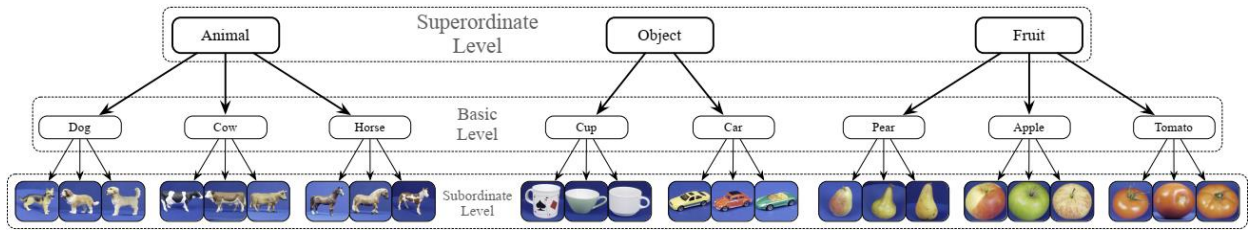

**Figure 3:** Example objects from the ETH-80 dataset along with the hierarchy of three categorization levels.

The CU3D-100 is organized into 100 categories with an average of 9{10 exemplars per category with controlled variability in pose and illumination. In this study, we report experimental results on the six categories of CU3D-100. Three categorization levels for the employed subset of CU3D-100 dataset are shown in Figure 4. The superordinate level includes the vehicle and non-vehicle categories. At the basic level, the vehicle category contains the car, airplane, and motorcycle; and the non-vehicle category includes the traffic cone, traffic light, and warning sign. The subordinate level includes four exemplars of each category at the basic level, resulting 6×4=24 categories at this level.

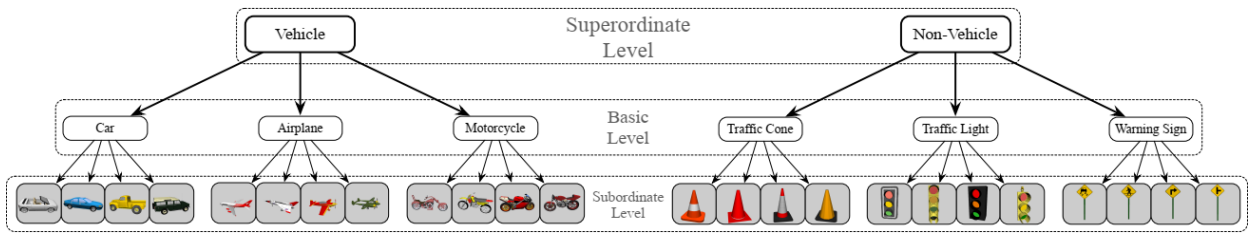

**Figure 4:** Example objects from the CU3D dataset along with the hierarchy of three categorization levels.

ImageNet is a dataset of over 15 million labeled high-resolution images belonging to roughly 22,000 categories [50]. Again, three categorization levels for the employed subset of ImageNet dataset are shown in Figure 5. The superordinate level includes the animal and non-animal categories. At the basic level, the animal category contains the bird and mammal; and the non-animal category includes the vehicle and ball. At the subordinate level, the bird category includes pigeon and duck; the mammal category includes cat and dog; the vehicle category includes bicycle and motorcycle; and the ball category includes soccer ball and tennis ball, resulting 4×2=8 categories at this level.

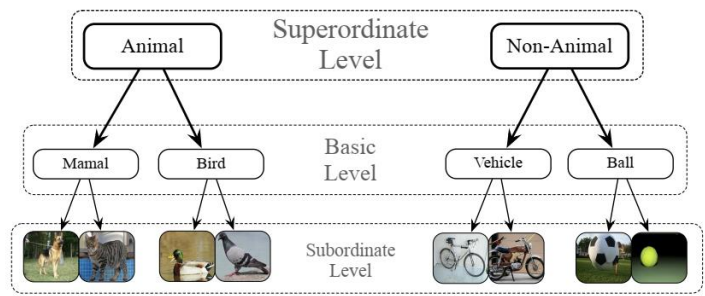

**Figure 5:** Example objects from the ImageNet dataset along with the hierarchy of three categorization levels.

It is worth mentioning that these datasets contain natural and unsegmented images, where objects have large variations in scale, viewpoint, and tilt, which makes their recognition hard [51].

## 3.2. Experimental setups

To perform the experiments, all the images in each dataset are filtered into low, intermediate, and high spatial frequency elements using appropriate Gabor filters. For all datasets, the images of each category are randomly divided into two separate sets with equal sizes for training and testing stages. For a fair comparison, identical training and testing sets are used in all compared models. Also, the statistical analysis of experimental results is accomplished with Analysis of Variance (ANOVA) test.

## 3.3. Experimental results

In the first experiment, the effect of applying different types of module (RSNN, SNN, CNN, DCNN, and HMAX) at each categorization level is investigated and the results are shown in Table 1. Note that in this experiment only different competitive models are examined at each categorization level. Input images of these modules are filtered into LSF, ISF, and HSF with respect to the categorization level. The result of using the R-SNN module, which its details is described in the proposed model, at each categorization level is shown in Table 1. Also, the result of using the SNN module at each categorization level is shown in this Table. The used SNN is trained with STDP, in an unsupervised manner, and three kinds of feature vectors of length n can be computed from the S2 layer: 1. The first-spike vector is a binary vector, in which all the values are zeros, except the one corresponding to the neuronal lattice with the earliest spike, 2. The spike-count vector stores the total number of emitted spikes by neurons in each lattice, 3. The potential vector contains the maximum membrane potential of the neurons in each lattice. After feature extraction for both training and testing data of each categorization level, the SVM classifier is used to evaluate the performance of each level. Additionally, at each categorization level, R-SNN is compared to a shallow CNN with a similar structure. The first convolutional layer of CNN has four feature maps with the same input window size as filters in the S1 of R-SNN. Then, a max-pooling layer is used with the same window size and stride of the neurons in the C1 of R-SNN. Next, a convolutional layer with n feature maps is used and the ReLU is set as the activation functions of convolutional neurons. Afterward, two dense layers: a hidden and output layer are used. In each task, number of neurons in the hidden layer is tuned. The ReLU and soft-max activation functions were employed for the hidden and output layers, respectively. To avoid overfitting, dropout regularization is performed on the dense layers and kernel regularization with different parameter values is applied. Also, at each categorization level, the R-SNN is compared with Alexnet that is a DCNN that significantly improved the recognition accuracy of the Imagenet dataset [20]. Alexnet is comprised of five convolutional layers followed by three fully connected layers. For our experiments, we have used the pre-trained version of Alexnet [52]. The SVM classifier is used to do the object recognition based on the extracted feature vectors from the 7th layer of pre-trained Alexnet. Also, the results of using HMAX module, which is one of the classic computational models of the object recognition process in visual cortex, at each categorization level is shown in Table 1.

As it is observed in Table 1, all models performed relatively good at the superordinate level, but at the basic and especially at the subordinate categorization levels their recognition rates decreased. The R-SNN module outperforms the SNN with different kinds of feature vectors at each categorization level. The R-SNN is a suitable choice for training class-specific neurons that are able to decide on the class of the input image using the first-spike latencies instead of an external classifier, but, the SNN is only a feature extractor unit and it requires an external classifier. So, not only the performance and biological plausibility of R-SNN is increased but also, its computational cost decreased. In addition, in most cases, the R-SNN works better than the shallow CNN, and Alexnet. Altogether, it seems that R-SNN is a good option for using in the categorization modules, based on the biological plausibility and also the efficiency of it relative to other state-of-the-art models.

**Table 1:** Comparison different models at different categorization levels on the ETH-80, CU3D, and ImageNet datasets.

| Dataset | Level | Model | | | | | | R-SNN |
|---|---|---|---|---|---|---|---|---|
| | | HMAX | CNN | Alexnet | SNN | | | |
| | | | | | First Spike | Spike Count | Max Potential | |
| ETH-80 | Superordinate | 96.4 | 97.2 | 96.8 | 95.1 | 94.9 | 95.8 | 97.4 |
| | Basic | 74.6 | 82.5 | 80.9 | 72.9 | 74.1 | 77.9 | 89.5 |
| | Subordinate | 57.9 | 64.8 | 63.7 | 56.4 | 59.3 | 59.8 | 68.7 |
| CU3D | Superordinate | 100 | 100 | 100 | 98.5 | 99 | 99 | 99.5 |
| | Basic | 93.8 | 95.9 | 94.3 | 93.9 | 94.3 | 94.7 | 96.4 |
| | Subordinate | 82 | 84.7 | 83.8 | 81.8 | 82.2 | 84.6 | 87.1 |
| ImageNet | Superordinate | 89.7 | 93.2 | 93.9 | 86.4 | 86 | 87.2 | 94.2 |
| | Basic | 73.5 | 80.1 | 78.7 | 74.3 | 74.3 | 74.6 | 77.4 |
| | Subordinate | 62.9 | 67.2 | 66.4 | 61.5 | 62.4 | 63.8 | 67.6 |

In the proposed model, the R-SNN module is utilized at each categorization level. To find suitable parameters' values an exhaustive search was performed at each categorization level of the used datasets. Table 2 outlines the parameters' values at each categorization level of the mentioned datasets.

**Table 2:** parameters' values of the R-SNN module at each categorization level for the ETH-80, CU3D, and ImageNet datasets.

| Dataset | Level | Parameters of R-SNN | | | | | | |
|---|---|---|---|---|---|---|---|---|
| | | Gabor Window Size | Number of Features | $M_r^+$ | $M_r^-$ | $M_p^+$ | $M_p^-$ | Threshold |
| ETH-80 | Superordinate | 27 | 24 | 0.025 | -0.025 | 0.01 | -0.005 | 155 |
| | Basic | 19 | 48 | 0.145 | -0.006 | 0.15 | -0.005 | 150 |
| | Subordinate | 13 | 336 | 0.27 | -0.025 | 0.055 | -0.0009 | 110 |
| CU3D | Superordinate | 27 | 16 | 0.047 | -0.025 | 0.01 | -0.005 | 135 |
| | Basic | 17 | 30 | 0.04 | -0.025 | 0.01 | -0.005 | 120 |
| | Subordinate | 11 | 192 | 0.16 | -0.025 | 0.01 | -0.001 | 110 |
| ImageNet | Superordinate | 29 | 120 | 0.14 | -0.04 | 0.01 | -0.005 | 120 |
| | Basic | 21 | 140 | 0.235 | -0.06 | 0.07 | -0.008 | 115 |
| | Subordinate | 15 | 176 | 0.6 | -0.102 | 0.01 | -0.001 | 110 |

The next experiment is carried out to investigate the influences of both spatial frequency and categorization level on the performance of SNN, R-SNN models for each mentioned dataset. The mean recognition accuracies of these models (averaged over 10 independent runs) with full-frequency bands of images (i.e., original images) as well as the frequency-filtered images (i.e., LSF, ISF, and HSF bands) for superordinate, basic, and subordinate categorization levels are shown in Figure 6. The bars denote the average recognition accuracies of each model, broken down by different spatial frequencies.

As it is shown in Figure 6, the categorization accuracies at the superordinate level are very high in all the frequency bands. This means that the coarse information at LSF bands is sufficient for categorization at the superordinate level. The average accuracies at the basic level are decreased with respect to those of the superordinate level and the maximum reduction is occurred in the LSF bands, while it is not changed much for the ISF, HSF, and full-bands. The same results can be observed for the subordinate level in which higher accuracy reduction occurs in the LSF and ISF bands. As shown, the LSF and ISF bands impair the categorization at the subordinate level relative to the former levels, meaning that higher frequency information is needed to perform categorization at subordinate level. In addition, the accuracies with the full-band images are higher than those with LSF, ISF, and HSF bands, but the ANOVA tests show that the differences are not significant between the full-band and the HSF bands (p>0.05) for all categorization levels. Altogether, instead of using the full-bands information on each level, only the corresponding and required spatial frequencies of the input image are sufficient.

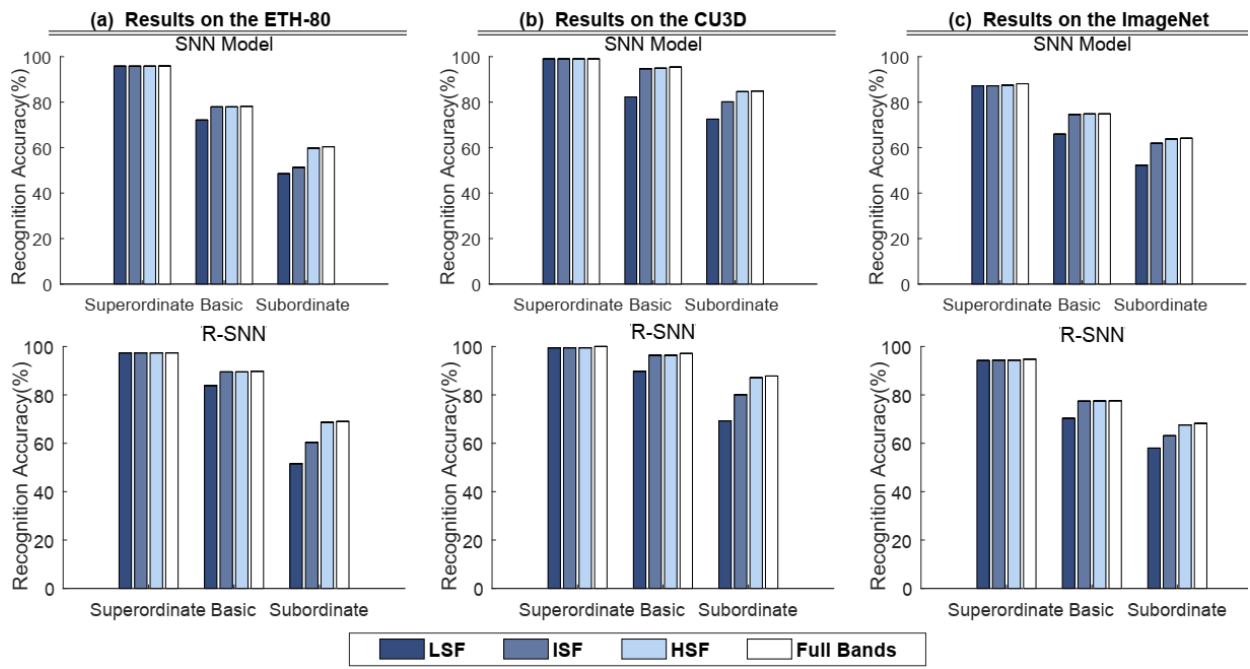

**Figure 6:** The Recognition accuracy of each model (SNN, R-SNN) at different categorization levels under different spatial frequency conditions. The results of models for the (a) ETH-80, (b) CU3D, and (c) ImageNet are represented in the first, second and third column, respectively.

The confusion matrices of R-SNN at the basic and subordinate level for ETH, CU3D, and ImageNet datasets are shown in Figure 7. As it is obvious, at the subordinate level the wrong decisions are increased with respect to the basic level, because the fine-grained categorization problem at this level. For example, at the subordinate level of the ETH-80 dataset, the R-SNN mistakes some of the Horse2 samples with Cow3 ones, which are resemble with together.

The next experiment focuses on the object recognition at different categorization levels in the case of partially occluded images. To this end, we prepared the partially occluded version of input images using a method similar to the Bubbles technique, in which some parts of images are occluded using uniformly positioned circular blob filters which are softened with Gaussian function [53]. Figure 8 shows the models' accuracies (SNN, R-SNN) at different categorization levels with the occluded images using the different number of circular blob filters (zero means no occlusion effect).

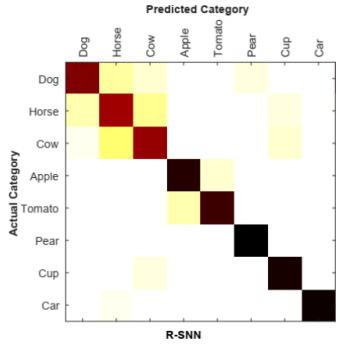
(a) Basic level of ETH-80 dataset

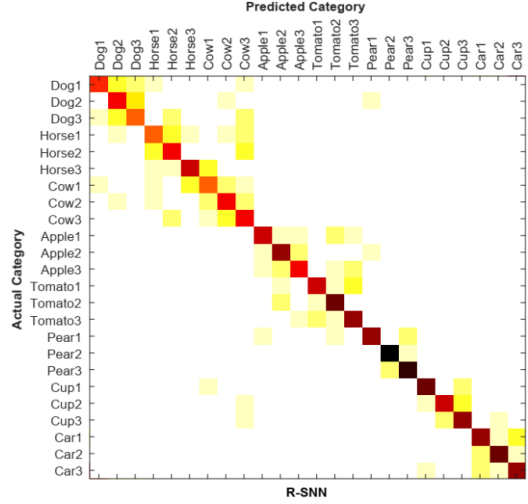
(b) Subordinate level of ETH-80 dataset

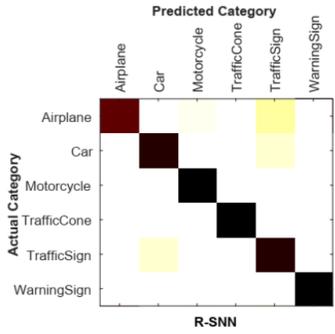
(c) Basic level of CU3D dataset

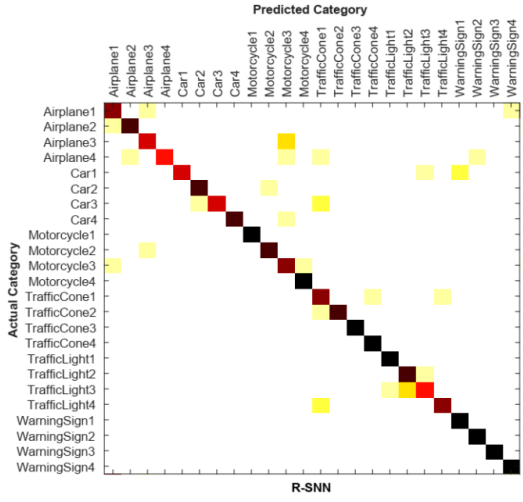
(d) Subordinate level of CU3D dataset

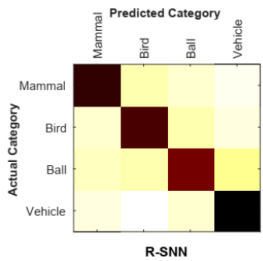
(e) Basic level of ImageNet dataset

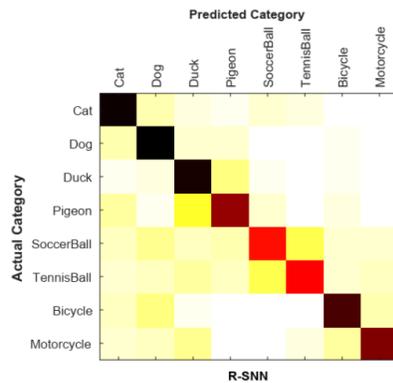
(f) Subordinate level of ImageNet dataset

**Figure 7:** The confusion matrices of R-SNN model at the a) basic and b) subordinate levels of ETH-80 dataset, c) basic and d) subordinate levels of CU3D dataset, and e) basic and f) subordinate levels of ImageNet dataset.

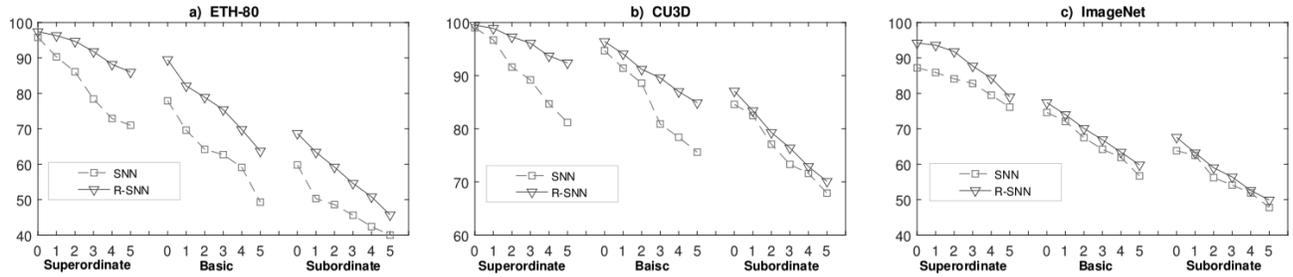

As can be observed in Figure 8, all models perform relatively good at the superordinate level, but at the basic and especially at the subordinate categorization levels their recognition rates decrease. Although the SNN is a biologically plausible model, its performance is very low in all cases. In all cases, R-SNN performs better than SNN due to using R-STDP. By increasing the number of blobs, the strength of occlusion is augmented and the recognition accuracies of these models are dropped in all categorization levels. In addition, the image degradation leads to relatively poor performance at the subordinate level, since it is thought that this level involves the most visual analysis of the input images. Thus, occlusion on the images leads to a strongly dropping in the recognition accuracies of the models at the subordinate level. In addition, the proposed model by considering the bio-inspired learning mechanism in its R-SNN module is more biologically acceptable than SNN. Also, the R-SNN uses a single module for both feature extraction and class specification of the input image, while the other models need a module comprising a feature extraction unit and then they use a classifier for specifying the class of the input images.

## 4. Conclusion and Future Perspectives

To date, various psychophysics experiments have been introduced for object recognition in different levels of abstraction, but constructing a computational model to mimic the object recognition in these categorization levels is neglected. In this paper, a bio-inspired computational model, is proposed for object recognition at different categorization levels. This hierarchical model contains superordinate, basic, and subordinate modules, in which each module comprised of R-SNN module for both feature extraction and classification phases. We suggest that LSF, ISF, and HSF elements of the input image can be used at the superordinate, basic, and subordinate levels, which can impressively decrease the computational load of our proposed model. To evaluate the proposed model, different experiments on the three well-known datasets are implemented.

The objects in the visual scene are not segregated from their own backgrounds such as images in the ImageNet dataset. Nevertheless, our visual system is extremely fast and accurate in recognizing objects in the natural scene. Furthermore, in addition to the variation in objects, their background is also changed, so extracting object-specific features is complicated. Confronting these defects, visual attention can be used in future research. It seems that visual attention would decrease the computational load of the model. Also, it would help to concentrate on the objects and its informative features to discriminate objects at each categorization level.

## Acknowledgement

This research received funding from Iran Cognitive Sciences and Technologies Council (No. 5859) and Iran National Science Foundation, INSF (No. 96002765).